\documentclass[runningheads]{llncs}

\usepackage{eccv}

\usepackage{eccvabbrv}

\usepackage{graphicx}
\usepackage{booktabs}

\usepackage{subcaption}
\usepackage{caption}
\usepackage{multirow}
\usepackage[table]{xcolor}

\usepackage[accsupp]{axessibility}  

\usepackage{hyperref}

\usepackage{orcidlink}

\usepackage{comment}

\newcommand{\method}{CCL}

\usepackage[utf8]{inputenc}

\usepackage{marvosym}

\begin{document}

\title{Improving Joint Audio-Video Generation with Cross-Modal Context Learning} 

\titlerunning{Cross-Modal Context Learning}

\author{Bingqi Ma \and Linlong Lang \and Ming Zhang \and Dailan He \and \\
 Xingtong Ge \and Yi Zhang \and Guanglu Song \and Yu Liu}

\authorrunning{B.~Ma et al.}

\institute{Vivix Group Limited \\
\email{mabingqi@vivix.ai\\
}
}

\maketitle

\begin{abstract}
The dual-stream transformer architecture-based joint audio-video generation method has become the dominant paradigm in current research. 
By incorporating pre-trained video diffusion models and audio diffusion models, along with a cross-modal interaction attention module, high-quality, temporally synchronized audio-video content can be generated with minimal training data.
In this paper, we first revisit the dual-stream transformer paradigm and further analyze its limitations, including model manifold variations caused by the gating mechanism controlling cross-modal interactions, biases in multi-modal background regions introduced by cross-modal attention, and the inconsistencies in multi-modal classifier-free guidance~(CFG) during training and inference, as well as conflicts between multiple conditions.
To alleviate these issues, we propose \textbf{Cross-Modal Context Learning}~(CCL), equipped with several carefully designed modules.
Temporally Aligned RoPE and Partitioning~(TARP) effectively enhances the temporal alignment between audio latent and video latent representations.
The Learnable Context Tokens~(LCT) and Dynamic Context Routing~(DCR) in the Cross-Modal Context Attention~(CCA) module provide stable unconditional anchors for cross-modal information, while dynamically routing based on different training tasks, further enhancing the model's convergence speed and generation quality.
During inference, Unconditional Context Guidance~(UCG) leverages the unconditional support provided by LCT to facilitate different forms of CFG, improving train-inference consistency and further alleviating conflicts.
Through comprehensive evaluations, \method~achieves state-of-the-art performance compared with recent academic methods while requiring substantially fewer resources.

\end{abstract}

\begin{figure}[t]
    \centering
    \includegraphics[width=0.95\linewidth]{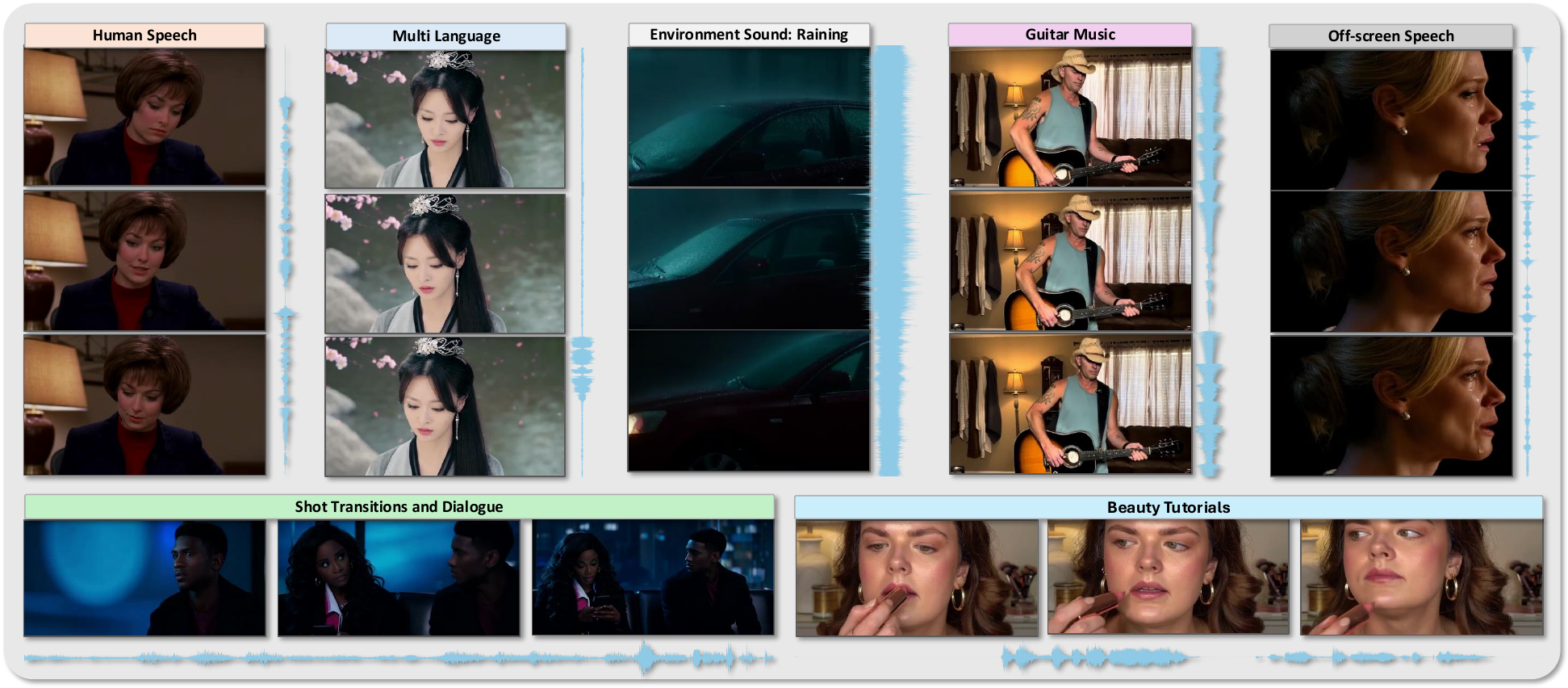}
    \caption{We demonstrate several capabilities of~\method, including multilingual human speech generation, environmental sound synthesis, music generation, background speech generation, storyboard-style scene transitions, and dialogue generation, as well as applicability to real-world scenarios such as beauty tutorial production.}
    \label{fig:showcase}
\end{figure}

\section{Introduction}
Diffusion models~\cite{ddpm,ddim} have achieved remarkable success in generative modeling for images~\cite{hunyuanimage,imagen,lidit,qwenimage,sd3,sdxl,dalle}, videos~\cite{hunyuanvideo,ltxvideo,opensora,sanavideo,wan,waver}, audio~\cite{audioldm,audioldm2,diffwave,makeanaoduio}, and text~\cite{diffusionllm,sahoo2024simple}.
Beyond single-modality synthesis~\cite{easyref, adt, neighborgrpo, swap, pretrain}, joint generation across multiple modalities has quickly emerged as a key research direction, aiming to produce coherent outputs that remain mutually consistent across modalities.
Recent studies~\cite{ovi, uniavgen, harmony, jova, klear, mmsonate, mavid} have placed particular emphasis on joint audio-video generation, demonstrating increasingly realistic synthesis and controllability.

With access to large-scale training corpora and growing model capacity, recent systems such as Veo3~\cite{veo}, Wan2.5~\cite{wan25}, Sora2~\cite{sora2}, Kling 3.0~\cite{kling3}, and Seedance 2.0~\cite{seedance2} exhibit strong prompt adherence, high visual fidelity, and temporally consistent motion.
Moreover, several open-source efforts, such as LTX-2~\cite{ltx2} and MOVA~\cite{movavideo}, also achieve strong performance, offering high generation quality and competitive audio-video coherence.
Meanwhile, academic studies~\cite{ovi, uniavgen, harmony, jova} on smaller-scale datasets have proposed new architectural designs and optimization strategies, providing deeper insights into the task and enabling a more systematic understanding of its challenges and opportunities.

Among these works, the dual-stream transformer architecture has been widely adopted.
Building on pre-trained video and audio diffusion models, cross-modal attention modules enable information exchange between the audio and video streams, thereby enabling synchronized audio and video content generation.
A representative design introduces an additional bidirectional cross-modal attention module into each diffusion transformer block to extract representation from another modality.
Some approaches\cite{jova,mmsonate,klear} attempt to facilitate cross-modal interaction between audio and video using a joint self-attention module, following the MMDiT\cite{sd3} paradigm.
However, these methods typically require training from scratch on billion-level audio-video pairs\cite{klear,mmsonate} or rely on auxiliary tasks designed specifically to ensure convergence\cite{jova}.
In this paper, we focus on the design of the bidirectional cross-attention paradigm, analyze key issues in existing designs, and propose corresponding solutions.

Existing methods typically use a gating mechanism to control the cross-modal interaction process.
Although this design does not introduce additional computational overhead, we find that switching the gate under different optimization objectives introduces a piecewise objective structure in parameter space, which destabilizes gradient directions across successive optimization steps and harms the convergence process, leading to limited generation performance.
Meanwhile, by analyzing the cross-modal attention map, we find that the model exhibits a strong bias towards regions in both the audio and video modalities that are unrelated to the other modality during cross-modal interaction.
This results in a negative impact on the convergence efficiency of the model's training and further leads to semantic confusion.
For the multi-modal CFG\cite{cfg}, existing methods often ignore the design of cross-modal unconditional information, or rely excessively on handcrafted priors, making it difficult to ensure alignment between training and inference.
There is also a conflict between text control and cross-modal control, which further affects the quality of the generated output.

To address the aforementioned issues, we propose a novel paradigm for joint audio-video generation, named \textbf{Cross-Modal Context Learning}.
The pipeline of~\method~is as shown in~\cref{fig:method}.
\method~inherits the traditional dual-stream transformer architecture, enhanced with the Temporally Aligned RoPE and Partitioning module, the Cross-Modal Context Attention module with Learnable Context Tokens and Dynamic Context Routing, as well as the Unconditional Context Guidance.
In TARP, we first apply temporal aligned Rotary Positional Embedding~(RoPE)~\cite{roformer} on the audio latent to address the temporal misalignment caused by differences in the audio and video sampling rates, as well as the VAE compression rates.
We further leverage a partitioning mechanism that restricts cross-modal attention between audio and video latents to a small temporal window, thereby simplifying the optimization of audiovisual consistency.
In each CCA module, we introduce LCT to provide contextual information from the other modality.
We expect these tokens to serve as a stable, averaged representation of the counterpart modality, enabling the cross-attention module to distinguish foreground from background information efficiently.
The DCR dynamically adjusts the composition of the keys and values involved in the computation of the CCA module.
During inference, UCG leverages the cross-modal context representation provided by LCT as unconditional information.
Such strategies encourage train-inference consistent, further enhancing the cross-modal coherence of the generated content without introducing adverse effects.

We also evaluate our proposed~\method~against several recent academic methods on the test set across multiple dimensions.
Experiments demonstrate that \method~ achieves competitive performance while requiring less training data and computation.
Extensive ablation studies further validate the effectiveness of each proposed module.
We provide some showcases in~\cref{fig:showcase} to further introduce the ability of \method~on multiple dimensions.

\section{Related work}

\subsection{Video Generation Models}

Video generation has undergone a paradigm shift driven by the rapid advancement of diffusion models.
Early video diffusion models rely heavily on U-Net architectures \cite{ronneberger2015u}. 
They expand pre-trained image generation models like Stable Video Diffusion \cite{ho2022video} and AnimateDiff~\cite{guo2023animatediff} by incorporating temporal attention mechanisms or 3D convolutions.
These methods effectively mitigate the shortage of high-quality video data. 
However, they face limitations when processing complex spatiotemporal dynamics and generating long videos.
Following the breakthrough of Sora~\cite{liu2024sora}, diffusion transformer (DiT) \cite{peebles2023scalable} gradually replace U-Net to become the dominant paradigm. 
The DiT architecture operates in the latent space and employs a 3D Variational Auto-Encoder~(VAE)\cite{vae} for efficient spatiotemporal compression. 
This design significantly improves the scaling capability of models on massive datasets and enhances the visual fidelity of generated videos.
Subsequently, numerous excellent open-source and closed-source models emerge, including the Wan~\cite{wan2025wan}, CogVideoX~\cite{yang2024cogvideox}, Hunyuan-Video~\cite{kong2024hunyuanvideo}, and Kling~\cite{wang2025kling}.
This architectural evolution and the effective utilization of large-scale training data establish the current state-of-the-art in video generation.

\subsection{Joint Audio-Video Generation}

Building upon the significant success of single-modality generation, cross-modal joint generation rapidly emerge as a frontier research topic in generative AI. 
Early attempts~\cite{omniavatar,wans2v} frequently rely on cascaded generation pipelines or entirely independent synthesis strategies.
Such disjointed frameworks inherently struggle to achieve precise temporal alignment and often produced mismatched semantic contexts.
To overcome these limitations, advanced close-source models like Veo3~\cite{veo} and Wan2.5~\cite{wan25} pioneer highly time-aligned joint audio-video generation.
Following this trend, recent open-source methods shift toward unified generation paradigms.
For instance, UniVerse-1\cite{wang2025universe} combines the Wan2.1 video backbone with the ACE-Step\cite{gong2025ace} audio model, utilizing cross-modal attention to align features for joint synthesis.
Ovi~\cite{ovi} employs a dual-branch structure that fuses pre-trained video and audio models, facilitating interaction through dedicated cross-attention modules.
Expanding on the dual-stream approach, LTX-2\cite{ltx2} introduces an asymmetric dual-stream transformer architecture, enforcing unified generation and strict temporal synchronization via cross-attention.

\section{Revisiting the Dual-Stream Transformer Pipeline}

\begin{figure}[t]
  \centering

  \begin{minipage}[t]{0.25\linewidth}
    \centering
    \includegraphics[width=\linewidth]{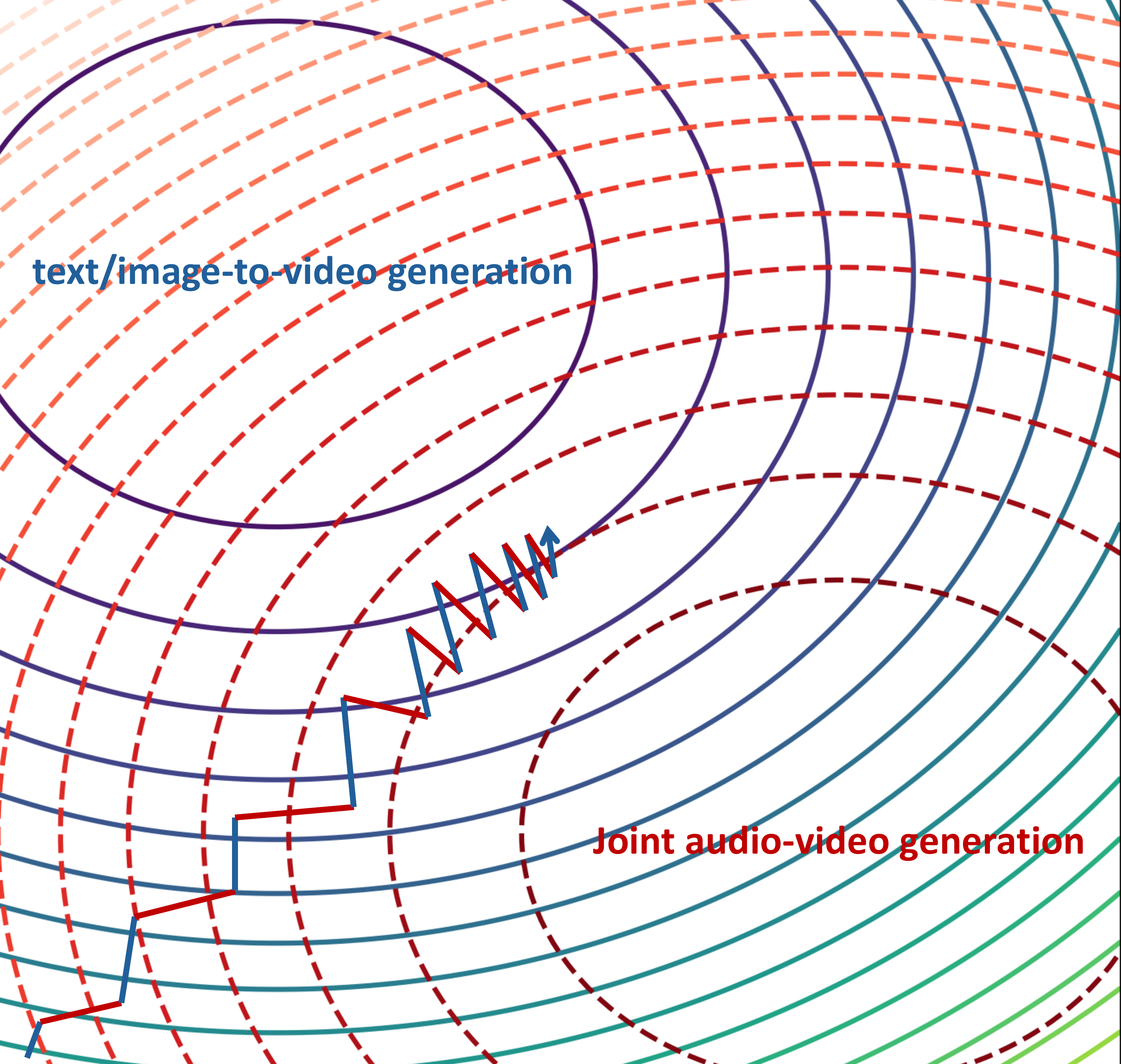}
    \captionsetup{font=scriptsize}
    \caption{The gating mechanism alters the optimization objective during training, which affects training efficiency.
    }
    \label{fig:figure1_1}
  \end{minipage}\hfill
  \begin{minipage}[t]{0.72\linewidth}
    \centering
    \includegraphics[width=\linewidth]{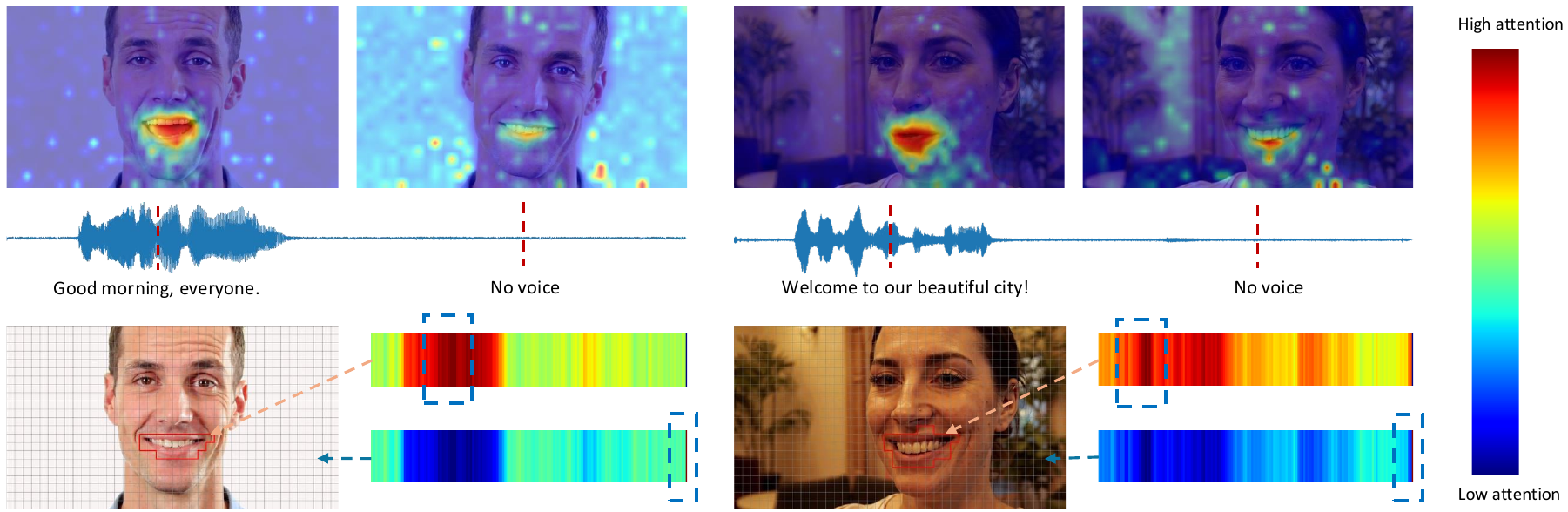}
    \captionsetup{font=scriptsize}
    \caption{The visualization of cross-modal attention in Ovi~\cite{ovi}. We observe that background regions in the audio attend strongly to random regions in the video, while background regions in the video assign high attention to the final audio token. This suggests that the cross-modal attention is semantically misaligned, introducing positional bias and consequently degrading model performance.}
    \label{fig:figure1_2}
  \end{minipage}\hfill
  
\end{figure}

In this section, we revisit the dual-stream transformer paradigm for joint audio–video generation and further investigate several design choices, analyzing the underlying principles and their limitations.

\subsection{The Gating Mechanism}
\label{sec:gate}
The gating mechanism provides an effective way to regulate cross-modal interactions while introducing no additional computational overhead.
We formulate a transformer block as follows:
\begin{equation}
\begin{aligned}
    \mathbf{X} &= \mathbf{X} + \mathrm{SA}(\mathbf{X}), \\
    \mathbf{X} &= \mathbf{X} + \mathrm{CA}(\mathbf{X}, \mathbf{T}), \\
    \mathbf{X} &= \mathbf{X} + g \cdot \mathrm{CA}(\mathbf{X}, \mathbf{Y}), \\
    \mathbf{X} &= \mathbf{X} + \mathrm{FFN}(\mathbf{X}),
\end{aligned}
\end{equation}
where $\mathbf{X}$ denotes the latent representation of the current modality; $\mathbf{T}$ is the text-conditioning embedding; and $\mathbf{Y}$ is the latent representation of the other modality. The scalar $g \in \{0, 1\}$ is a binary gate that controls whether features from the other modality are injected.
For single-modality generation, we simply set $g=0$; for joint audio-video interaction, we set $g=1$.

However, we argue that the gating mechanism will significantly harm the convergence of the optimization process.
Under different optimization objectives, changes in the gate state can induce a piecewise structure of the objective in parameter space, making the gradient direction across successive optimization steps difficult to stabilize.
\cref{fig:figure1_1} provides a clearer illustration of this phenomenon, and this conclusion is also confirmed in~\cref{sec:ablation}.

\subsection{Cross-Modal Attention}
\label{sec:attention}
Bidirectional cross-attention is a key module for establishing associations between the two modalities.
Attention maps can reveal the attention correlations between the two modalities.
For example, in audio-driven talking-head scenarios, the audio modality typically exhibits closer attention over the speaker’s mouth region, enabling lip movements to align with the spoken content.

However, there is a local correlation between the audio and video modalities. Specifically, the mouth region occupies only a small foreground area in the video, and most of the background video representation does not exhibit a direct correlation with the audio representation.
Therefore, this cross-attention module needs to simultaneously distinguish between foreground and background while correctly driving the lip movements.
A similar local structure is also present in the audio representation.
In~\cref{fig:figure1_2}, we show some visualization cases generated by Ovi~\cite{ovi}.
We observe that background video tokens show high attention values with the last audio token, while background audio tokens exhibit strong attention at random spatial locations in the image.
This consistency across multiple test cases suggests that, for background regions, the model has learned a positional bias that is independent of the semantic content.
This means that the model typically requires extensive training to converge, and suffer from semantic confusion.

\subsection{Multi-Modal CFG}
The multi-modal CFG decouples the control strength of the text and cross-modal representations, allowing dynamic adjustment based on different requirements, thereby enabling flexible generation control.
We can formulate the multi-modal CFG as follows:
\begin{equation}
\resizebox{0.9\linewidth}{!}{$
\hat v_\theta(x_t, t)
=
v_\theta\!\left(x_t, t \mid {uc}_{\text{text}}, \varnothing\right)
+
s_{\text{text}}
\Big(
v_\theta\!\left(x_t, t \mid c_{\text{text}}, \varnothing\right)
-
v_\theta\!\left(x_t, t \mid {uc}_{\text{text}}, \varnothing\right)
\Big)
+
s_{\text{m}}
\Big(
v_\theta\!\left(x_t, t \mid {uc}_{\text{text}}, c_{\text{m}}\right)
-
v_\theta\!\left(x_t, t \mid {uc}_{\text{text}}, \varnothing\right)
\Big),
$}
\end{equation}
where $x_t$ denotes the latent sample at inference timestep $t$,
$v_\theta(x_t, t \mid c_{\text{text}}, c_{\text{m}})$ is the flow conditioned on the text $c_{\text{text}}$ and the cross-modal condition $c_{\text{m}}$, $uc_{text}$ is the negative prompt, $s_{\text{text}}$ and $s_{\text{m}}$ are the CFG scales.
LTX-2 designs a variant of the multi-modal CFG, which further enhances performance based on conditional information.
However, the underlying approach remains consistent, so we do not provide a detailed explanation.

In text-conditioned visual generation tasks, unconditional information plays a crucial role. 
By setting well-defined negative prompts, the quality and stability of the generated results can be effectively improved across various dimensions.
However, existing joint audio-video generation methods typically construct unconditional information by dropping the modality information, such as setting the cross-modal attention gate to $0$.
Harmony\cite{harmony} presents an initial attempt that performs inference by using a static video and silent audio as the unconditional cross-modal representation.
However, such unconditional information does not appear in the training distribution and relies excessively on handcrafted priors, resulting in a mismatch between training and inference.
This design also requires an additional forward pass to obtain the unconditional representation, incurring extra cost.
Meanwhile, MOVA\cite{movavideo} also observes that the multi-modal CFG can lead to conflicts between text control and cross-modal control. When a larger cross-modal CFG scale is set, better audio-video coherence is achieved, but the capacity of the audio stream is significantly compromised.

\section{Cross-Modal Context Learning}

\begin{figure}[t]
  \centering
  \includegraphics[width=\linewidth]{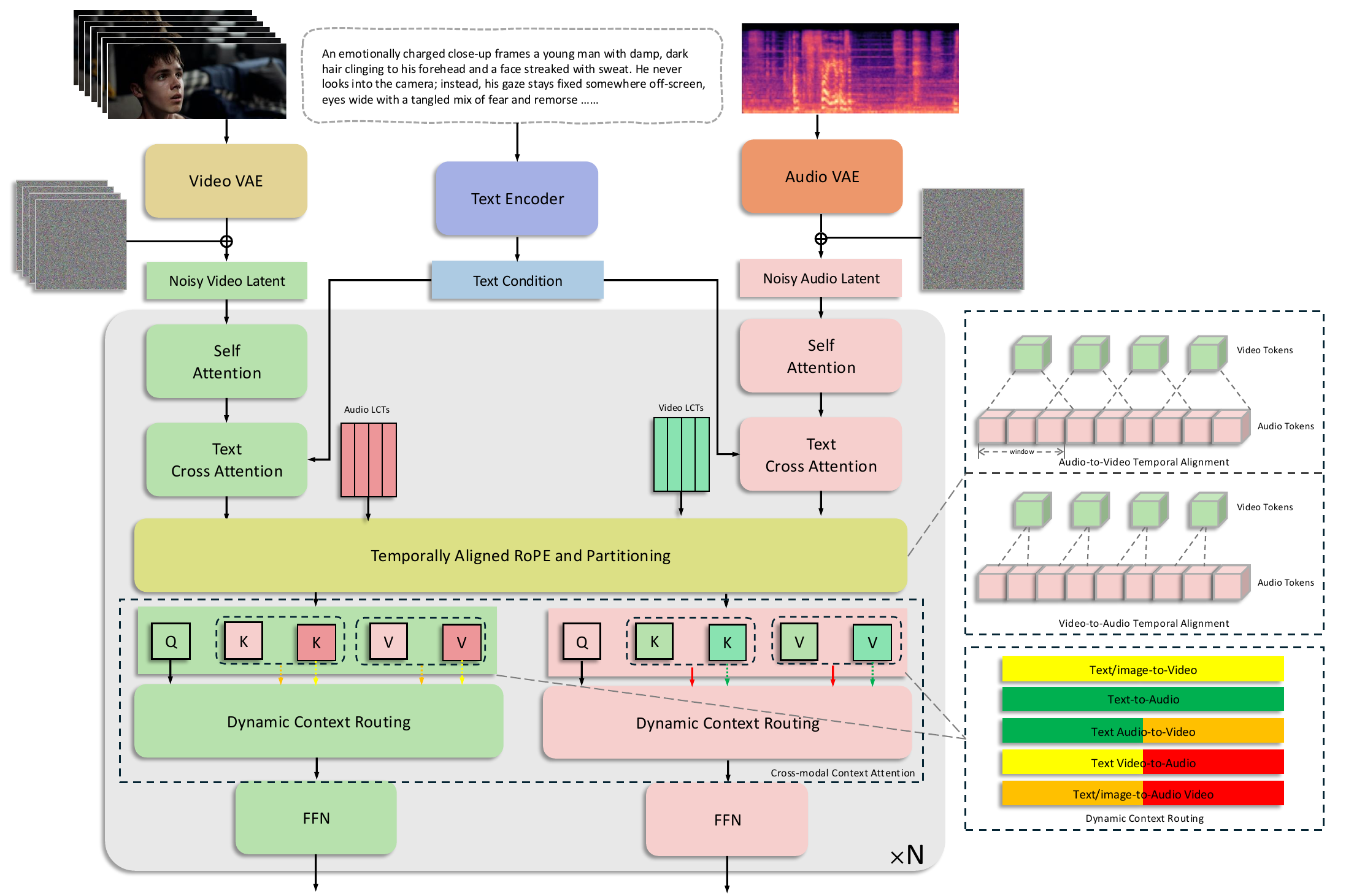}
  \caption{The pipeline of our proposed \textbf{Cross-Modal Context Learning}. CCL follows the conventional dual-stream transformer architecture, equipped with several novel-designed modules, enabling efficient and effective joint audio-video generation with high consistency. The figure illustrates the implementation details of proposed modules. For Dynamic Context Routing, the various colors denote that the corresponding colored paths on the left are in an activated state.}
  \label{fig:method}
\end{figure}
The pipeline of~\method~is as shown in~\cref{fig:method}.
The video frames are compressed into the latent space via a 3D VAE. 
Raw audio is first converted into a Mel-spectrogram and then compressed into a 1D latent representation using an audio VAE.
For both the audio and video modalities, we use a unified prompt format, where the spoken content is enclosed in quotation marks.
Both the video and audio streams are built on transformer blocks and share the same layer number and overall architecture, except that the audio branch uses fewer channels to improve parameter efficiency.
Each transformer block consists of a self-attention module, a text-conditioned cross-attention module, a Temporally Aligned RoPE and Partitioning module, a Cross-Modal Context Attention module, and a feed-forward network layer.
In the following sections, we will introduce the design of core modules.

We define several variables to make the following presentation clearer.
Let $\mathbf{X}_a \in \mathbb{R}^{b \times t_a \times d_a}$ denote the audio latent in an audio transformer block output by the text-conditioned cross-attention module, where $b$ is the batch size, $t_a$ is the number of audio tokens, and $d_a$ is the dimension of the audio transformer block.
Similarly, $\mathbf{X}_v \in \mathbb{R}^{b \times s_v \times d_v}$ denotes the video latent, where $s_v = t_v \cdot h \cdot w $ is the number of video tokens, with $(t_v, h, w)$ being the latent resolution produced by the 3D VAE, and $d_v$ is the dimension of the video transformer block.
Due to the mismatch between $d_a$ and $d_v$, the QKV projections in the CCA module should align the feature dimensions across modalities, yielding $\mathbf{q}_a \in \mathbb{R}^{b \times t_a \times d_a}$, $\mathbf{k}_a, \mathbf{v}_a \in \mathbb{R}^{b \times t_a \times d_v}$, $\mathbf{q}_v \in \mathbb{R}^{b \times s_v \times d_v}$, and $\mathbf{k}_v, \mathbf{v}_v \in \mathbb{R}^{b \times s_v \times d_a}$.

\subsection{Temporally Aligned RoPE and Partitioning}
\label{sec:align}
Prior works~\cite{harmony,uniavgen} have shown that temporally aligning audio and video latent can substantially accelerate training convergence and improve audio–video consistency.
Our TARP module adopts a similar design.
In \cref{fig:method}, we also visualize the specific operations.

Due to the mismatch between $t_a$ and $t_v$, directly adopting standard RoPE on both audio and video latent results in a misalignment between positional embeddings and temporal position.
For $\mathbf{q}_a$ and $\mathbf{k}_a$, we rescale the indices of RoPE by a factor of \(\tfrac{t_v}{t_a}\), yielding $\bar{\mathbf{q}}_a$ and $\bar{\mathbf{k}}_a$. 
For $\mathbf{q}_v$ and $\mathbf{k}_v$, we apply the standard RoPE, yielding $\bar{\mathbf{q}}_v$ and $\bar{\mathbf{k}}_v$.

In addition to the temporally aligned RoPE, we further leverage a partitioning mechanism to simplify the optimization of audio-video consistency.
Each modality does not need to attend to the other modality over the entire temporal span, but only a small local neighborhood around the current timestep.
Let $c=\left\lfloor \frac{t_a}{t_v} \right\rfloor$ be the number of strictly aligned audio latent tokens per video latent frame.
For the $i$-th video frame, the window center index is
$m_i=\bigl\lfloor \tfrac{c}{2}+c\cdot i \bigr\rfloor$, and we set the window size $s=3\cdot c$.
We reshape $\bar{\mathbf{q}}_v$ into $\tilde{\mathbf{q}}_v \in \mathbb{R}^{(b\cdot t_v)\times (h\cdot  w)\times d_v}$, and partition $\bar{\mathbf{k}}_a$ and $\mathbf{v}_a$ to capture richer contextual information and improve temporal continuity, yielding $\tilde{\mathbf{k}}_a,\,\tilde{\mathbf{v}}_a \in \mathbb{R}^{(b \cdot t_v)\times s \times d_v}$.
Then $\tilde{\mathbf{q}}_v$, $\tilde{\mathbf{k}}_a$ and $\tilde{\mathbf{v}}_a$ will be fed into the audio-to-video CCA module.
A similar strategy is adopted on $\bar{\mathbf{q}}_a$, $\bar{\mathbf{k}}_v$, and $\bar{\mathbf{v}}_v$.
Considering the audio information exhibits weaker temporal sensitivity to the video information, we set the window size $s=1$ and directly adopt the nearest-neighbor interpolation.
$\tilde{\mathbf{q}}_a \in \mathbb{R}^{(b\cdot t_a) \times 1 \times d_a}$ and $\tilde{\mathbf{k}}_v, \tilde{\mathbf{v}}_v \in \mathbb{R}^{(b \cdot t_a) \times (h\cdot w) \times d_a}$ will be fed into the video-to-audio CCA module.

\subsection{Cross-Modal Context Attention}
The CCA module is built upon the conventional bidirectional cross-modal cross-attention module by further introducing Learnable Context Tokens and Dynamic Context Routing.
\subsubsection{Learnable Context Tokens}
We introduce Learnable Context Tokens~(LCT) into each CCA module to provide contextual information from the other modality.
We expect these tokens to serve as a stable, averaged representation of the counterpart modality.
Such stability provides an anchor for tokens in the current modality that are less relevant to cross-modal interactions, enabling the cross-attention module to distinguish foreground from background information efficiently.
This paradigm effectively accelerates convergence and improves the quality of joint generation.

We use $\mathbf{L}_{\text{a}} \in \mathbb{R}^{n_a \times d_v}$, where $n_a$ is the token number, to represent the LCT in each audio-to-video CCA module.
Symmetrically, we use $\mathbf{L}_{\text{v}} \in \mathbb{R}^{n_v \times d_a}$, where $n_v$ is the token number, to represent the LCT in each video-to-audio CCA module.
Subsequently, we apply KV projections on LCT and repeat them in the batch dimension to align with keys and values from each modality, obtaining $\mathbf{k}^l_a, \mathbf{v}^l_a \in \mathbb{R}^{(b\cdot t_v) \times n_a \times d_v}$ and $\mathbf{k}^l_v, \mathbf{v}^l_v \in \mathbb{R}^{(b\cdot t_a) \times n_v \times d_a}$.
Specifically, we do not adopt RoPE on LCT due to its temporal-insensitive nature.
We describe the computation procedure using joint audio-video generation as an example.
Subsequently, we performed the concatenation operation and obtained:
\begin{equation}
\begin{aligned}
\langle\hat{\mathbf{k}}_a &= \text{concat}(\tilde{\mathbf{k}}_a, \mathbf{k}^l_a), \quad \hat{\mathbf{v}}_a = \text{concat}(\tilde{\mathbf{v}}_a, \mathbf{v}^l_a)\rangle \in \mathbb{R}^{(b \cdot t_v) \times (s + n_a) \times d_v} \\
\langle\hat{\mathbf{k}}_v &= \text{concat}(\tilde{\mathbf{k}}_v, \mathbf{k}^l_v), \quad \hat{\mathbf{v}}_v = \text{concat}(\tilde{\mathbf{v}}_v, \mathbf{v}^l_v)\rangle \in \mathbb{R}^{(b \cdot t_a) \times (h \cdot w + n_v) \times d_a}.
\end{aligned}
\end{equation}
Finally, we conduct the cross-attention computation:
\begin{equation}
\mathbf{O}_a = \text{attention}\left(\bar{\mathbf{q}}_a, \hat{\mathbf{k}}_v, \hat{\mathbf{v}}_v\right), \mathbf{O}_v = \text{attention}\left(\bar{\mathbf{q}}_v, \hat{\mathbf{k}}_a, \hat{\mathbf{v}}_a\right),
\end{equation}
then reshape back to the original shape like $\mathbf{X}_a$ and $\mathbf{X}_v$.
Since $n_a$ and $n_v$ are far smaller than the latent token number, this design incurs only negligible additional computational overhead.

\subsubsection{Dynamic Context Routing}
We propose the Dynamic Context Routing to address the substantial slowdown in convergence and inferior performance of the gating mechanism. 
Dynamic Context Routing can dynamically adjust the composition of the keys and values involved in the computation of the CCA module.
This design encourages the model to learn more consistent contextual representations in multi-task training settings, thereby significantly accelerating convergence while further improving the model's representational capacity.

The training objectives of \method~include text/image-to-video, text-to-audio, audio-to-video, video-to-audio, and joint audio-video generation.
When training the text/image-to-video and text-to-audio tasks, the CCA module interacts directly with the corresponding LCT, ignoring the cross-modal latent representations.
The behaviors of the two streams are independent and do not affect each other.
When training the audio-to-video generation task, the timestep of the audio stream is set to 0.
In the audio stream, the CCA module interacts only with the corresponding video LCT, ignoring the video latent representations.
In the video stream, the CCA module interacts with both the audio LCT and the audio latent representations.
When training the video-to-audio generation task, the timestep of the video stream is set to 0.
In the video stream, the CCA module interacts only with the corresponding audio LCT, ignoring the audio latent representations.
In the audio stream, the CCA module interacts with both the video LCT and the video latent representations.
For the joint audio-video training task, we synchronize the timesteps of the two streams, and each CCA module interacts with both the corresponding LCT and the information from the other modality.
To better illustrate the execution of DCR, we provide a visualization in \cref{fig:method}.
Different colors denote modality-selection patterns in DCR.
For the text/image-to-video task, only the LCT within each block participate in cross-modal attention, and thus only the yellow pathway is active.
For the text-audio-to-video task, the green and orange pathways are active.
For joint audio-video generation, all pathways are active.

\subsection{Unconditional Context Guidance}
In \method, the LCT are designed to provide stable anchor points for background information, meaning that LCT become natural unconditional representations after training.
Compared to manually designed silent audio or static videos, LCT exhibit stronger consistency between training and inference.
At the same time, using LCT as unconditional representations eliminates the need for additional inference, further enhancing efficiency.
For different inference costs, we provide two forms of UCG. 
If we aim to obtain the generation result through two inference steps, we use the following form:
\begin{equation}
\resizebox{0.9\linewidth}{!}{$
\hat v_\theta(x_t, t)
=
v_\theta\!\left(x_t, t \mid {uc}_{\text{text}}, \mathcal{L}^{m}\right)
+
s_{m}
\Big(
v_\theta\!\left(x_t, t \mid c_{\text{text}}, c_{\text{m}}\right)
-
v_\theta\!\left(x_t, t \mid {uc}_{\text{text}}, \mathcal{L}^{m}\right)
\Big),
$}
\end{equation}
where $s_m$ is the CFG scale for current modality and $\mathcal{L}^{m}$ represents the LCT for current unconditional modality.
If we hope to decouple the control strengths of text and cross-modal representations, we can define it in conjunction with multi-modal CFG as follows:
\begin{equation}
\resizebox{0.9\linewidth}{!}{$
\hat v_\theta(x_t, t)
=
v_\theta\!\left(x_t, t \mid {uc}_{\text{text}}, \mathcal{L}^{m}\right)
+
s_{\text{text}}
\Big(
v_\theta\!\left(x_t, t \mid c_{\text{text}}, \mathcal{L}^{m}\right)
-
v_\theta\!\left(x_t, t \mid {uc}_{\text{text}}, \mathcal{L}^{m}\right)
\Big)
+
s_{\text{m}}
\Big(
v_\theta\!\left(x_t, t \mid {uc}_{\text{text}}, c_{\text{m}}\right)
-
v_\theta\!\left(x_t, t \mid {uc}_{\text{text}}, \mathcal{L}^{m}\right)
\Big).
$}
\end{equation}
In this case, we need three steps to complete a single inference.
UCG also effectively alleviates the conflict between text condition and cross-modal conditions, especially the video condition in the audio stream.
Traditional modality drop methods provide cross-modal CFG with a strong indication of whether a modality is present, which is reasonable for the video stream receiving audio conditions. 
However, the audio stream does not require such a strong direction; rather, it benefits from the enhancement of audio-related information within the video stream.
In this case, LCT plays the role of a robust condition guider, preventing issues caused by overly strong cross-modal conditions, such as increased error rates in the audio modality.

\subsection{Multi-Task Training}
During training, we assign each task a distinct sampling probability and optimize only the loss from the generative objectives.
We optimize~\method~using flow matching\cite{flowmatching} as the training objective.
We formulate the multi-task training as follows:
\begin{equation}
\scalebox{0.9}{%
$\begin{aligned}
\mathcal{L}=
\begin{cases}
\mathcal{L}_{\mathrm{aud}}+\mathcal{L}_{\mathrm{vid}}, & \text{text/image-to-video, text-to-audio, or joint audio-video} \\
0\cdot \mathcal{L}_{\mathrm{aud}}+\mathcal{L}_{\mathrm{vid}}, & \text{audio-to-video} \\
\mathcal{L}_{\mathrm{aud}}+0\cdot \mathcal{L}_{\mathrm{vid}}, & \text{video-to-audio}
\end{cases}
\end{aligned}$%
}
\end{equation}
where $\mathcal{L}_{\mathrm{aud}}$ and $\mathcal{L}_{\mathrm{vid}}$ denote the training losses for the audio and video streams, respectively.
Notably, during audio-to-video and video-to-audio training, we detach the gradients of the reference stream to prevent it from being over-optimized when its timestep is set to 0, which could otherwise induce abnormal behavior.

\begin{figure}[t]
  \centering
  \includegraphics[width=\linewidth]{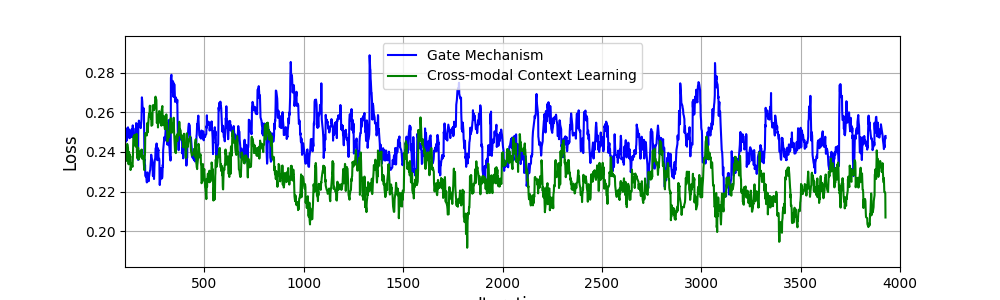}
  \caption{The visualization of the training loss when adopting the gate mechanism compared with leveraging the~\method. We only sampled the loss for the joint audio-video generation task and applied the EMA operation. Notably, due to the instability of the loss during the early stages of training, we begin visualizing from iteration 100.}
  \label{fig:loss}
\end{figure}

\begin{figure}[t]
  \centering
  \includegraphics[width=\linewidth]{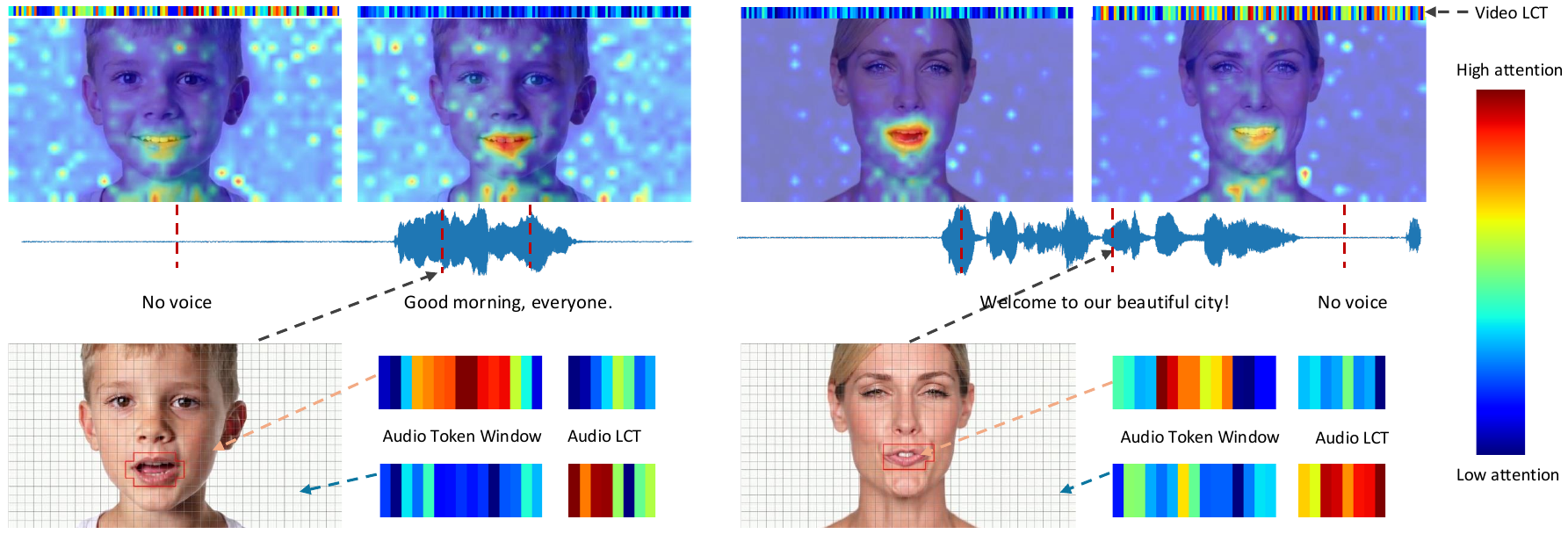}
  \caption{The visualization of attention maps in~\method. We observe that background tokens in both the audio and video modalities attend more strongly to the LCT.}
  \label{fig:ccl_attention}
\end{figure}

\section{Visualizing the Training Process and Attention Map}
To better understand~\method, we provide visual analyses of both the training dynamics and the learned attention behavior. 
Specifically, we visualize the loss curves throughout the training process, as well as the attention maps of the trained model.
As shown in~\cref{fig:loss}, we find that \method~exhibits a substantially faster and more stable decrease in training loss compared to the baseline approach that adopts the gating mechanism. 
This trend suggests that \method~facilitates more efficient optimization, enabling the model to reach a lower-loss regime earlier during training. The quantitative results of both experiments are summarized in~\cref{tab:ablation1}, which further corroborate our observations.

For the attention map visualization in~\cref{fig:ccl_attention}, we observe that background tokens in both audio and video modalities tend to attend more strongly to the LCT. 
This implies that \method~encourages the model to capture modality-specific background patterns and to form a clearer separation between foreground and background content. 
Such a separation can reduce the ambiguity in cross-modal alignment and alleviate the optimization difficulty, leading to lower training loss and faster convergence.

\section{Experiments}
\subsection{Experimental Settings}
\subsubsection{Datasets and Training}
We have curated a dataset consisting of million-level audio-video pairs for joint training.  
This dataset includes the widely-used OpenHumanVid~\cite{openhumanvid} as well as additional in-house collections.  
It encompasses a diverse range of sources, such as interviews, short dramas, and films.  
For the pretraining of the audio diffusion model, we also incorporate academic audio datasets, including Wavcaps~\cite{wavcaps} and VGGSound~\cite{vggsound}.  
The video diffusion stream is initialized with the parameters from Wan2.1-14B.  
The audio diffusion stream follows the same architecture but smaller channel number for efficiency.
To facilitate multimodal interaction, we integrate a CCA module at each block of the network.  
The model training is performed in two stages: first, the audio diffusion pretraining, followed by joint audio-video training.  

During the pretraining phase of the audio diffusion model, we use a learning rate of $1 \times 10^{-4}$ and train for 160,000 steps with a batch size of 3200.  
In the joint audio-video training phase, we increase the learning rate for the cross-modal attention parameters to $1 \times 10^{-4}$, while keeping the learning rate for all other parameters at $1 \times 10^{-5}$.
During the ablation phase, we trained for 6000 steps with a batch size of 128.  
In the final setting, we used a batch size of 320 and trained for 12,000 steps.
The audio LCT number $n_a$ is set as 8, and the video LCT number $n_v$ is set as 128.
In the multi-task training phase, the probabilities for text/image-to-audio/video, audio-to-video, video-to-audio, and joint audio-video generation are set to 0.1, 0.15, 0.15, and 0.6, respectively.  
The entire training process is conducted at a resolution of $352\times640\text{p}$.

\subsubsection{Evaluation Metrics}
To ensure a fair comparison, we constructed a 200-sample test set with no overlap with the training data. 
Each test sample consists of an image and a corresponding prompt that describes both the video content and the audio content. 
The test set is divided into the easy subset and the hard subset.
In the easy subset, the test images are primarily portraits and upper-body shots, and the prompts specify that the person remains static. 
In the hard subset, the test images feature full-body views, and the prompts include body movements. The two subsets share the same audio content.
During evaluation, we also refined the prompt to accommodate different models' preferences for input structure.

We primarily consider three dimensions: audio quality, lip-sync, and AV-Alignment.
For audio quality, we adopt the Meta AudioBox Aesthetics~\cite{aesbox} to obtain production quality~(PQ) and content usefulness~(CU).
We also use the Whisper-large-v3~\cite{whisper} model to compute the word error rate (WER) of the generated audio.
For the lip-sync metric, we use SyncNet~\cite{syncnet} to compute Sync-C and Sync-D for each generated audio-video pair.
For the AV-align metric, we use the ImageBind score~(IB-Score)~\cite{imagebind} computed on joint audio-video embeddings, as well as the DeSync score predicted by Synchformer~\cite{synchformer}.
We report performance on easy and hard test sets separately for both Lip Sync and AV-Alignment. 
For Audio Quality, we report the mean score across the two subsets.

\begin{table}[t]
    \centering
    \caption{We compare \method~with existing methods on the test set, including Ovi~\cite{ovi}, LTX-2~\cite{ltx2}, and MOVA~\cite{movavideo}. The evaluation metrics cover audio quality, lip sync, and AV-Alignment. The results show that \method~substantially outperforms prior approaches, further validating the effectiveness of our proposed method.}
    \resizebox{\textwidth}{!}{%
    {%
    \setlength{\tabcolsep}{6pt}%
    \renewcommand{\arraystretch}{1.1}%
    \begin{tabular}{l c | c c c | c c | c c | c c | c c}
    \toprule[1.5pt]
    \multirow{2}{*}{Method} & \multirow{2}{*}{Data}
      & \multicolumn{3}{|c|}{Audio Quality}
      & \multicolumn{2}{|c|}{Lip Sync-easy}
      & \multicolumn{2}{|c|}{Lip Sync-hard}
      & \multicolumn{2}{|c|}{AV-Alignment-easy}
      & \multicolumn{2}{|c}{AV-Alignment-hard} \\
    \cline{3-13}
      &  & CU$\uparrow$ & PQ$\uparrow$ & WER$\downarrow$
      & Sync-C$\uparrow$ & Sync-D$\downarrow$
      & Sync-C$\uparrow$ & Sync-D$\downarrow$
      & DeSync$\downarrow$ & IB$\uparrow$
      & DeSync$\downarrow$ & IB$\uparrow$ \\
    \hline
    Ovi\cite{ovi}                     & 30.7M & 5.88 & 6.14 & 17.51\% & \textbf{8.47} & 7.32 & 5.96 & 8.57 & \textbf{1.11} & 0.20 & \textbf{1.18} & 0.21 \\
    LTX-2\cite{ltx2}                  & -- & 5.12 & 5.35 & 11.04\% & 7.67 & 7.90 & 5.47 & 8.93 & 1.18 & 0.19 & 1.20 & 0.21 \\
    MOVA\cite{movavideo}              & 50M & \textbf{6.34} & \textbf{7.24} & 8.09\% & 7.79 & 7.70 & 5.38 & 8.87 & 1.15 & 0.20 & 1.22 & 0.19 \\
    \hline
    \rowcolor{gray!20}\textbf{CCL}    & 4M & 5.71 & 6.42 & \textbf{4.62}\%  & 8.45 & \textbf{7.48} & \textbf{6.12} & \textbf{8.25} & \textbf{1.11} & \textbf{0.23} & 1.19 & \textbf{0.22} \\
    \bottomrule[1.5pt]
    \end{tabular}
    }%
    }
    \label{tab:method}
\end{table}

\subsection{Comparison with Previous Works}
We compare our proposed \method~with several representative prior works.
In particular, we focus on recent open-source models for joint audio generation, including Ovi~\cite{ovi} as well as the more recent LTX-2~\cite{ltx2} and MOVA~\cite{movavideo}.
These methods all adopt a similar dual-stream transformer architecture.
We perform inference on the test set using the official scripting setup without any modifications.
As shown in~\cref{tab:method}, compared with recent works such as LTX-2 and MOVA, Ovi still maintains competitive performance.
This observation is consistent with the conclusions reported in the MOVA paper.
Notably, the multi-modal CFG in MOVA leads to a pronounced degradation in audio quality.
Therefore, following the official codebase configuration, we do not use multi-modal CFG during inference.
In terms of audio quality, MOVA achieves the best performance on the CU and PQ metrics.
Ovi and \method~attain comparable results on these two metrics.
However, on WER, \method~exhibits a substantial advantage, outperforming MOVA by 3.47\%.
On the lip-sync metric, \method~also demonstrates superior performance compared with Ovi.
Except for the Sync-C metric on the easy test set, \method~achieves state-of-the-art performance across all other metrics.
On the AV-Alignment metric, \method~also achieves leading performance.
This further demonstrates the effectiveness of our proposed approach.

\subsection{Ablation Study}
\label{sec:ablation}
We conduct a detailed ablation study of the proposed modules.
For audio quality, we primarily report WER.
For lip-sync and AV alignment, we take the mean over the easy and hard subsets for simplicity.
The baseline setting uses traditional dual-stream architecture similar to Ovi~\cite{ovi} and adopts the gating mechanism for multi-task training.
At inference time, we follow the same strategy as Ovi, where the cross-modal representation is used in both conditional and unconditional inference, and the guidance is applied only on the text condition.
As shown in~\cref{tab:ablation1}, introducing TARP improves convergence efficiency; for instance, Sync-C increases from 4.53 to 5.05, while Sync-D decreases from 10.81 to 10.17.
After replacing the gating mechanism with LCT and DCR, the model achieves further and substantial gains, suggesting that the background-anchored representations provided by cross-modal context can ease optimization and improve training efficiency.
With UCG further introduced, we observe clear improvements in lip-sync and AV alignment; for example, Sync-C increases by 0.52, Sync-D decreases by 1.03, DeSync decreases by 0.02, and IB increases by 0.01.
MOVA~\cite{movavideo} reports that multi-modal CFG can increase WER by a large margin; in contrast, UCG retains the benefits of multi-modal CFG while avoiding this issue.
We also integrate the static-unconditional design from SyncCFG~\cite{harmony} into UCG for verification, but observe a slight drop across all metrics.
This suggests that UCG is not only more compute-efficient, but also yields better overall performance than SyncCFG.

\begin{table}[t]
\centering
\caption{We conduct extensive ablation studies to validate the effectiveness of each proposed module.
We also compare our design choices with existing alternatives featuring similar components.
The results indicate that every introduced module contributes to consistent performance improvements.}
\resizebox{0.8\textwidth}{!}{%
{%
\setlength{\tabcolsep}{6pt}%
\renewcommand{\arraystretch}{1.1}%
\begin{tabular}{c c c c | c | c c | c c}
\toprule[1.5pt]
\multirow{2}{*}{TARP} & \multirow{2}{*}{LCT/DCR} & \multirow{2}{*}{UCG} & \multirow{2}{*}{SyncCFG~\cite{harmony}}
  & \multicolumn{1}{c|}{Audio Quality}
  & \multicolumn{2}{c|}{Lip Sync}
  & \multicolumn{2}{c}{AV-Alignment} \\
\cline{5-9}
  & & & 
  & WER$\downarrow$
  & Sync-C$\uparrow$ & Sync-D$\downarrow$
  & DeSync$\downarrow$ & IB$\uparrow$ \\
\hline
  &  &  &                                                          & \textbf{4.52}\% & 4.53 & 10.81 & 1.20 & 0.20 \\ 
$\checkmark$ &  &  &                                               & 4.67\% & 5.05 & 10.17  & 1.20 & 0.20 \\
$\checkmark$ & $\checkmark$  &  &                                  & 4.81\% & 5.62 & 9.12  & 1.18 & 0.21 \\
\rowcolor{gray!20}$\checkmark$ & $\checkmark$  & $\checkmark$ &    & 4.63\% & \textbf{6.14} & \textbf{8.09}  & \textbf{1.16} & \textbf{0.22} \\
$\checkmark$ & $\checkmark$  & $\checkmark$ & $\checkmark$         & 5.59\% & 5.85 & 8.46  & 1.18 & 0.20 \\
\bottomrule[1.5pt]
\end{tabular}
}%
}
\label{tab:ablation1}
\end{table}

\section{Conclusions}
In this paper, we revisit the dual-stream transformer paradigm and examine key limitations, including manifold variations introduced by gating-based cross-modal control, biases in background regions from cross-modal attention, and inconsistencies between training and inference in classifier-free guidance with conflicting conditions.
To address these issues, we propose Cross-Modal Context Learning, which integrates several tailored modules. 
Temporally Aligned RoPE and Partitioning module improves audio–video temporal alignment. Learnable Context Tokens and Dynamic Context Routing within Cross-Modal Context Attention provide stable anchors and task-adaptive routing, improving convergence and generation quality. 
During inference, Unconditional Context Guidance leverages unconditional support from LCTs to better match training behavior and mitigate conflicts between text and cross-modal conditions.
Extensive experiments show that \method~achieves state-of-the-art performance against recent methods while using substantially fewer computational resources.

\newpage

%
%

\end{document}